\documentclass{article}
\usepackage{graphics}
\usepackage{subcaption} 
\usepackage{acra}
\usepackage{afterpage}
\usepackage{graphicx}
\usepackage{amsmath,amssymb}
\usepackage{makecell}
\usepackage{algorithmicx}
\usepackage{algorithm} 
\usepackage[noend]{algpseudocode}
\usepackage{appendix}
\usepackage{xcolor}
\usepackage{dblfloatfix}

\makeatletter
\def\BState{\State\hskip-\ALG@thistlm}
\makeatother

\title{Bounded Exploration with World Model Uncertainty in Soft Actor-Critic Reinforcement Learning Algorithm}
\author{Ting QIAO, Henry Williams, David Valencia, Bruce MacDonald\\ 
Centre for Automation and Robotic Engineering Science\\ 
The University of Auckland, New Zealand.\\
tqia574@aucklanduni.ac.nz, henry.williams@auckland.ac.nz}

\begin{document}
\maketitle
\begin{abstract}
One of the bottlenecks preventing Deep Reinforcement Learning algorithms (DRL) from real-world applications is how to explore the environment and collect informative transitions efficiently. The present paper describes \emph{bounded exploration}, a novel exploration method that integrates both `soft' and intrinsic motivation exploration. Bounded exploration notably improved the Soft Actor-Critic algorithm's performance and its model-based extension's converging speed. It achieved the highest score in $6/8$ experiments. Bounded exploration presents an alternative method to introduce intrinsic motivations to exploration when the original reward function has strict meanings.  
\end{abstract}

\section{Introduction}
$f_{\theta}(x_{t+1},r_{t}|s_t, a_t)$
    Reinforcement learning refers to a class of autonomous learning methods that learn from trial and error. Among them, Model-Free Reinforcement Learning (MFRL) provides a flexible and easy-to-implement way to control autonomous systems like robots. Thus, a robot is capable of tackling hard-to-engineer tasks. However, their data inefficiency becomes a significant obstacle preventing MFRL from solving complex real-world problems \cite{kober2013reinforcement}. Training an MFRL agent may demand millions of interactions with an environment to learn control skills \cite{kober2013reinforcement}. However, the mechanical parts in a robot can quickly wear and tear. How to actively explore the uncertain region in an environment and avoid repetition is crucial to this issue \cite{ladosz2022exploration}. 

    In terms of exploration, there are primarily two possible routes. From one perspective, \cite{ziebart2010modeling} has studied the relationship between a policy's entropy and rewards. Ziebart's work has been extended to state-of-art `soft' MFRL algorithms \cite{levine2018reinforcement}. Soft RL introduces a policy entropy term into the optimization objective. A soft agent acts to maximize future rewards and expand their action distribution to encourage exploration. Alternatively, other research quantified information theoretic measures such as Mutual Information as bonuses to encourage exploration \cite{ladosz2022exploration}. The existing literature primarily considers two methods separately. Little attention has been given to how these two methods can be integrated.

\begin{figure}[ht]
    \centering
    \includegraphics[width=\linewidth]{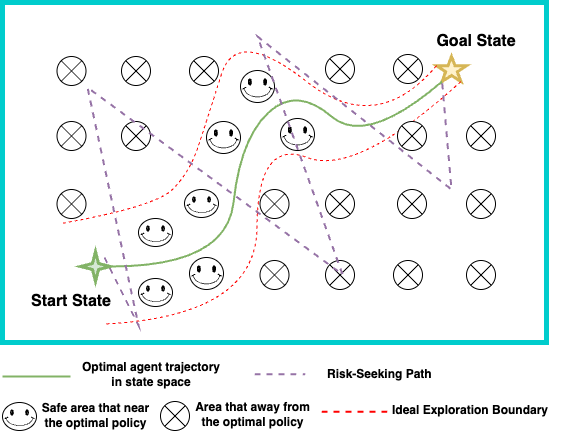}
    \caption{An oracle agent can follow the optimal path \color{olive}(Green line)\color{black}\ that always has the highest rewards. In reality, an agent needs to find this path by trial-and-error, presumably in the \color{red}(Smile faces within red dot lines)\color{black}\ region. However, the agent can also pursue exploration bonuses ($\otimes$), resulting in a  \color{purple} purple trajectory\color{black}\ , which can be far away from the optimal path.}
    \label{fig:mesh0}
    \end{figure}

    Recently, world models in Model-Based Reinforcement Learning (MBRL) have been extended to determine to what extent the agent is uncertain about the future states \cite{latyshev2023intrinsic}. Research from Barto \cite{barto2013intrinsic} suggests that an agent should learn to pursue this uncertainty as an `intrinsic motivation' that emerges from the agent itself. Conventionally, uncertainty is added to a reward function to encourage exploration \cite{pathak2017curiosity,aubret2019survey}. However, if the mingle of rewards and bonuses is not carefully designed (Figure \ref{fig:mesh0}), an agent may exploit exploration bonuses leading to risk-seeking behaviour \cite{garcia2015comprehensive} and high regret \cite{auer2002finite}. A robot controlled by a novelty-exploitation agent may walk into hazard states \cite{brunke2022safe}.  

    This work hypothesizes that an agent shall exploit uncertainty within the area suggested by a soft algorithm. In other words, an agent may explore under the guidance of the original reward function. The present research studied a Soft Actor-Critic (SAC) \cite{haarnoja2018soft} as an example. We called the proposed exploration strategy \emph{bounded exploration}, a novel intrinsic-motivation-pursuing approach that does not change the original reward function. The bounded exploration algorithm combines two types of exploration strategies by selecting highly uncertain actions within a soft policy's distribution. In the present research, a vital assumption is that an agent wants to explore uncertain areas to improve learning efficiency but within the bounds of quality actions.

\section{Background}
    Markov Decision Process (MDP) describes a system in which future states only depend on the current state and the action an agent is about to take. It is formulated by its state space $\mathcal{S}$, action space $\mathcal{A}$, a transition model $p(s_{t+1}|s_t,a_t): \mathcal{S} \times \mathcal{S} \times \mathcal{A} \rightarrow [0,1]$ and a reward function $r(s_t, a_t):\mathcal{S}\times\mathcal{A}\rightarrow[r_{min},r_{max}]$. $r_{max}$ is suggested to be negative to penalise control costs. Reinforcement Learning algorithms are designed to find the optimal policy $\pi^{*}$ that maximizes the expectation of future discounted returns $\mathbb{E}[V(s_0)]=\mathbb{E}[\sum_{t=0}^{T}\gamma^{t} r_{t}]$. In the recent decade, function approximators ($\pi_{\theta}(a_t|s_t)$), such as neural networks, have been used to approximate such an optimal policy. Among them, a probabilistic neural network policy outputting a distribution of actions offers more flexibility.  
    
    \subsection{Soft Actor-Critic}
        Soft algorithms formulate a policy's entropy and reward maximization into one objective (Equation \ref{soft_objective}). It can automatically balance exploration and exploitation. 
        
        \begin{equation}\label{soft_objective}
            \pi^{*} = \arg \max_{\pi} \sum_{t} \mathbb{E}_{(s_t,a_t)\sim\rho_{\pi}}[r(s_t, a_t) + \alpha \mathcal{H}(\pi(.|s_t))].
        \end{equation}
        
\begin{figure}[ht]
        \centering
        \includegraphics[width=0.45\textwidth]{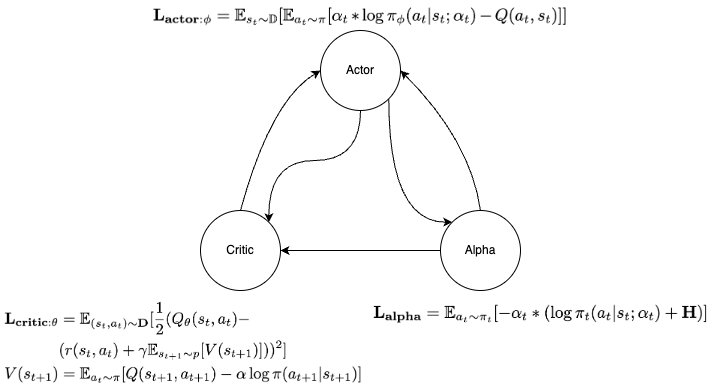}
        \caption{Soft Actor-Critic Updating Map ($\mathbf{L}$ is a loss function)}
        \label{fig:mesh1}
        \end{figure}

        In soft algorithms, Soft Actor-Critic (SAC) takes advantage of the Actor-Critic framework that allows the policy to output continuous actions. The reward $r(s_t,a_t)$ is interpreted as ``the log of optimality probability $\log p(\mathcal{O}|s_t)$", where $p(\mathcal{O}|s_t)$ is the probability of taking the optimal action at state $s_t$. A temperature coefficient $\alpha_t$ is added to control updating magnitude. Figure \ref{fig:mesh1} shows how to update each part. 
        
        As the updating map in Figure \ref{fig:mesh1} shows, the Q value and temperature $\alpha_t$ decide the loss function of the actor. The actor is updated to maximize the Q value and minimize the negative entropy $\mathbb{E}_{a_t\sim\pi}[\log\pi_{\phi}(a_t,s_t;\alpha_t)]$ that limits exploration. Even so, an agent can only output one action at each time step. This action should carry as much information as possible to the environment. 
    
\section{Related Works}
 
    \begin{figure*}[ht]
    \centering
    \includegraphics[width=\textwidth]{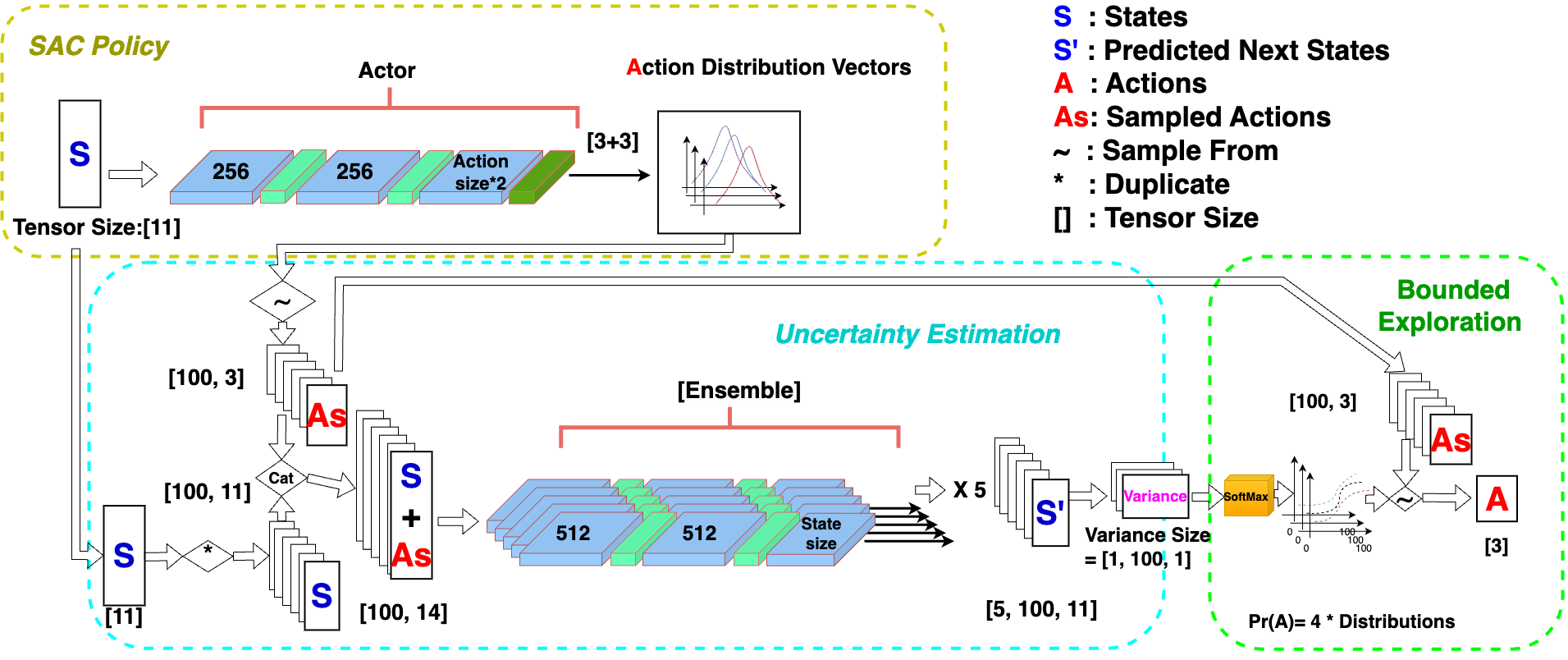}
    \caption{For each current state do: Step 1 (SAC Policy): the soft policy parameterize Normal distributions that $\mathbf{N}$ action candidates (e.g. $\mathbf{N}=100$) are sampled from, Step 2 (Uncertainty Estimation): feed-forward action candidates and states to an ensemble of world models (e.g. $\Omega_{\mathbf{M}=5}$) to compute uncertainty, Step 3 (Bounded Exploration): select the action causing the highest world-model uncertainty(selecting one action to execute from the $\mathbf{N}=100$).}
    \label{fig:mesh2}
    \end{figure*}


    There is a considerable amount of literature \cite{latyshev2023intrinsic} on employing world-model uncertainty for exploration. Among them, a significant portion of strategies require quantifying uncertainty. Bounded exploration falls into this category. Early work uses a Kullback–Leibler divergence to measure the predictive uncertainty \cite{houthooft2016vime} as an intrinsic motivation. Recent research suggests that aleatoric uncertainty is more suitable for this purpose \cite{depeweg2018decomposition}. Pathak et al. \cite{pathak2017curiosity} have developed an Intrinsic Curiosity Module (ICM) that uses an inverse model to compute the prediction error in action space as their intrinsic motivation. It has avoided the paradox of exploring to get novel states or getting novel states and then knowing where to explore. Despite the encouraging results from previous research, nearly all add exploration bonuses to the original rewards after quantification. In real-world applications, rewards may have specific meanings. For instance, a reward can correspond to the inverse distance to a goal. Adding an uncertainty measure to a distance could distort the meaning of the reward.  
    
    More recent research has attempted to utilise uncertainty in different ways. \cite{lee2021sunrise} quantifies the uncertainty with the variance of a set of Q-value estimators and adds the variances to Temporal-Difference updates. They attribute the credit of its performance gain to the re-weighting of target Q-values based on uncertainty estimation. \cite{chua2018deep} uses random-shooting rollouts to utilise the diverse synthetic data that an ensemble of neural networks can generate. Their method received nearly twice as many rewards as their Soft Actor-Critic baseline. \cite{buckman2018sample} has developed the Stochastic Ensemble Value Expansion (STEVE) algorithm that discounts transitions along the Q-value expansion horizon in Model-based Value Expansion (MVE) \cite{feinberg2018model} based on the inverse of model uncertainty. STEVE can dis-encourage highly uncertain transitions, which is likely to cause catastrophic failures \cite{luo2018algorithmic,feinberg2018model} in MBRL. However, an integration of 'soft' exploration and intrinsic-motivation-based exploration has been overlooked.
    
    From an exploration perspective, executing high-uncertainty actions is preferred. \cite{yao2021sample}'s research shares the same point of view. Their optimal actions are selected as $a_t = \arg\max_{a_t}(Q(s_t,a_t) + \hat{V}(s_t, a_t))$, where the $\hat{V}$ is the uncertainty estimation. However, their work has not isolated the effect of introducing uncertainty in action selection. Besides, introducing Q-values may also encourage the agent to take 'on-policy' actions, leading to a higher reward result. Moreover, consistently selecting the highest value actions may ignore other actions, potentially leading to sub-optimal results. 

\section{Methods}
    
    Bounded Exploration allows world-model uncertainty to decide what action $a_t$ is novel for the agent to explore from a set of SAC proposed actions. An agent conducting bounded exploration makes decisions of actions to transmit information of both `where SAC wants to go' and `where the world model is uncertain about'. Bounded exploration was instantiated by selecting novel actions from candidates bounded by the policy distribution. The hypothesis is deliberately selecting such actions leads to safer and more efficient exploration than adding an uncertainty bonus to the reward function. One of the reasons is that the agent will concentrate on possible actions suggested by the reward function rather than exploiting exploration bonuses.  
    
    The present method is detailed in Figure \ref{fig:mesh2}. The novel work is illustrated in the bounded exploration part in the green rectangle. This section will elaborate on preparing candidate actions first, then obtaining uncertainty estimation, and how to use it to participate in decision-making. 

    \subsection{Prepare action candidates}
\begin{figure}[ht]
    \centering
    \includegraphics[width=0.45\textwidth]{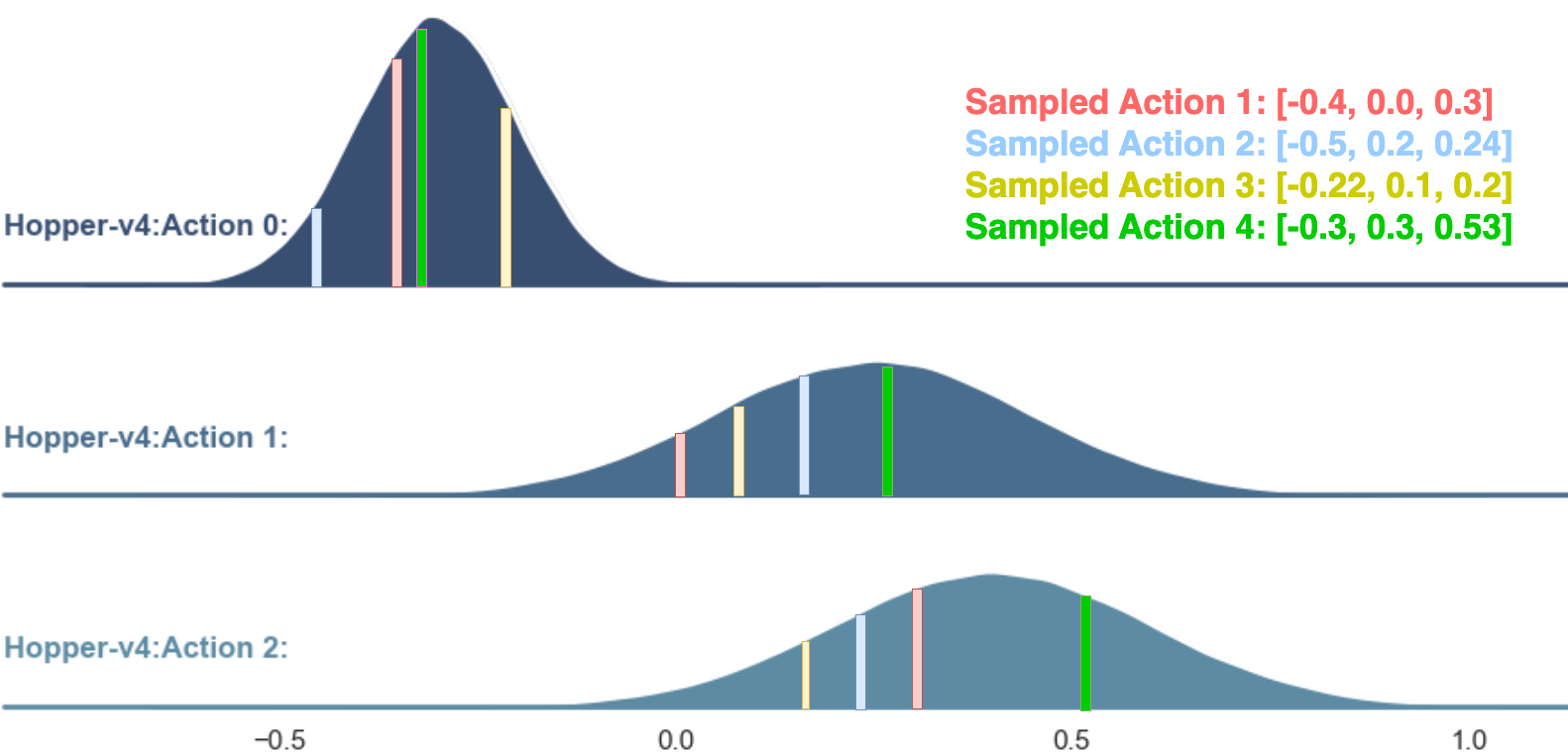}
    \caption{Sample Actions from a Multi-variant Stochastic Policy (e.g. Hopper-v4): $\mathbf{N}=4$ samples in different colours were drawn from the distribution.}
    \label{fig:mesh3}
    \end{figure}

    The present paper employed a SAC algorithm with auto-tuning temperature as the framework. As illustrated in Figure \ref{fig:mesh3}, a SAC agent can supply a distribution of actions $\pi_{\theta}(a_{t}|x_t)$ rather than numerical values. These distributions have already indicated the exploration-exploitation competence based on the received rewards. One of the benefits of having a distribution of actions is that it can be sampled several times. Each sample represents part of the information from the SAC. In Figure \ref{fig:mesh3}, these policy distributions were sampled $\mathbf{N}$ (e.g. $\mathbf{N}=4$) times as state-action pair candidates $\{(s_t,a_t^{n})\}_{n=0}^{N=4}$.
    
    \subsection{Uncertainty Estimation}
        
    Capturing world models' uncertainty as an agent's intrinsic motivation has been widely investigated and received promising results \cite{depeweg2016learning,chua2018deep,kurutach2018model,janner2019trust}. Compared with other intrinsic motivations, such as prediction error, inferencing uncertainty does not need target labels.
    
    A lack of training data can cause epistemic uncertainty \cite{kendall2017uncertainties}. It has been suggested \cite{depeweg2018decomposition} that such epistemic uncertainty is a viable option for intrinsic motivation. It needs an ensemble of world models to identify. In an ensemble, different world models trained with different data collaborate as a group of `experts' from different backgrounds. They may predict different results with the same input. Therefore, epistemic uncertainty could be quantified via models' discrepancy \cite{pathak2019self}.  
    
    We opted to use the variance of predictions from each model as an uncertainty measure for simplicity. The present research formulated an ensemble of neural networks as $\Omega = \{p_{\phi}^{m}(s_{t+1}|s_{t}, a_{t})\}_{m=1}^{M}$, where $m$ is the index of each neural network, $\phi$ is a neural network's parameters. Assuming an ensemble of world models $\Omega$ contains 5 neural networks $\Omega_{M=5}$, for a single state-action pair $(s_t, a_t^{n})$, the ensemble outputs 5 predictions $\{s_{t+1}^{m}\}_{m=1}^{M=5}$. The variance of $\{s_{t+1}^{m}\}_{m=1}^{M=5}$ in each dimension $\sigma^{2}_{d}$ is worked out first. Then, the variance in each dimension $d$ was summed up as the uncertainty measure:
    \begin{equation}\label{uncertainty_each}
        u_n = \sum_{d=1}^{D} \sigma_{d}^{2}
    \end{equation}
    Each uncertainty measure $u_n$ was assigned to one state-action pair $(s_t,a_t^{n})$. 

    The ensemble of world models in this research was trained to predict normalised state differences $\Delta=s_{t+1}-s_{t}$, reminiscent of the training target in \cite{nagabandi2018neural}\cite{kurutach2018model}. A random model was selected when the ensemble was needed to make a prediction. Regarding the model-based part, a short horizon length $H=2$ was used as suggested in \cite{janner2019trust}. It is worth noting that the present research fed the same training data to both the world models and the SAC agent.


    \subsection{Select one action with high uncertainty}
    
    The novelty of bounded exploration comes from using uncertainty to select actions for exploration (Green box: Bounded Exploration in Figure \ref{fig:mesh2}). From the agent's perspective, it only has one chance (action) to interact with the environment at each time step. An action should convey as much information as possible about where the agent wants to explore. Bounded exploration aims to include information from both the SAC agent and world models. With this in mind, a rank-preserving Gibbs distribution (Equation \ref{gibbs_distribution}) was built over candidate actions $\{a_t^{n}\}_{n=1}^{n=N}$. The probability of taking each candidate action relied on the uncertainty measure $u_n$ from forwarding the state-action pair $(s_t,a_t^{n})$ to the ensemble of world models $\Omega$. The probability of taking candidate $i$ is given as:
    \begin{equation}\label{gibbs_distribution}
        p(a_t^i) = \frac{\exp(u_n)}{\sum_{n=1}^{N}\exp(u_n)}
    \end{equation}

    During training, it was noticed that the difference in uncertainty between different actions is not apparent. The reason could be that the action dimension is significantly smaller than the state dimension. A normalization was applied before computing the probability. The result distribution was sampled $S$ times. The action closest to the mean of the SAC policy's distribution was outputted. It represented the `decision' from the SAC side and brought the out-of-SAC-distribution actions back. Bounded exploration viewed $S$ as a parameter. If $S$ is high, the agent chooses actions close to the original SAC actions rather than exploratory actions.
    
    The proposed method represented a valuable alternative to utilising intrinsic motivations. Compared with Yao's \cite{yao2021sample} method that uses $a_t = \arg\max_{a_t}(Q(s_t,a_t) + \hat{V}(s_t, a_t))$, bounded exploration did not use Q value. Therefore, it does not prefer on-policy actions. This method successfully combined SAC exploration and world-model intrinsic exploration.

    \begin{figure}[ht]
    \centering
    \includegraphics[width=0.48\textwidth]{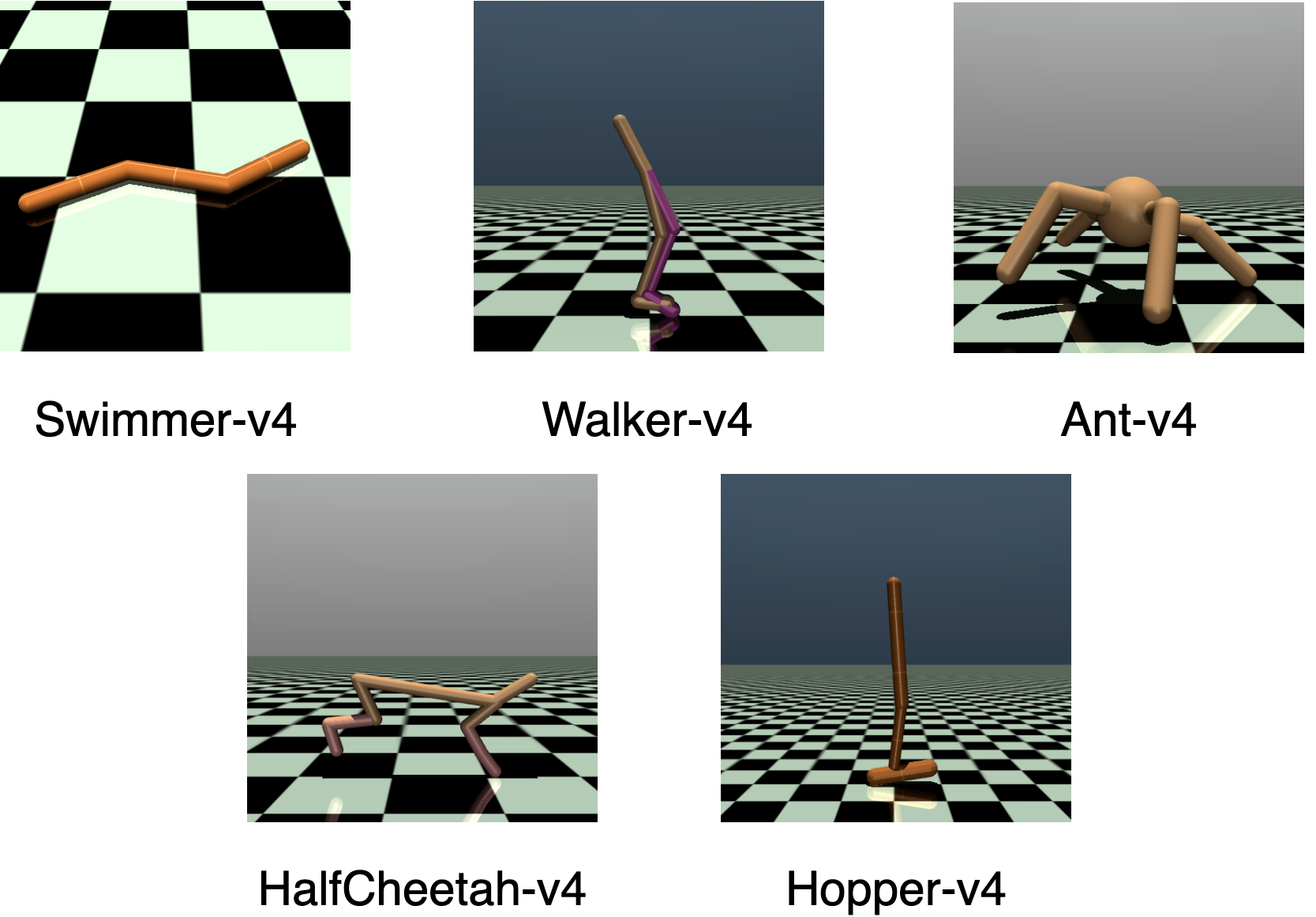}
    \caption{Mujoco Environments.}
    \label{fig:mesh4}
    \end{figure}
    
    \begin{table*}[!b]
        \centering
        \caption{Averaged performance over the last 10 evaluation results. Bold scores indicate the highest result. (SAC: Vanilla Soft Actor Critic. BE: Bounded Exploration. QU: Q-value and Uncertainty)}
        \label{tab:performance-all}
        \begin{tabular}{ccccccc}
        \hline
        \textbf{Method} & \textbf{HalfCheetah} & \textbf{Swimmer} & \textbf{Ant}      & \textbf{Hopper}     & \textbf{Walker2d} \\
        \hline
        SAC             & 9200.0 $\pm$ 2258.5           & 68.2$\pm$7.7           & -151.5$\pm$251.8             & \textbf{776.1 $\pm$242.8}            & 172.8 $\pm$157.4\\ 
        \textbf{SAC+BE} & \textbf{9747.7 $\pm$ 2444.5} & \textbf{140.4$\pm$9.9} & \textbf{-102.1$\pm$156.2}  & 475.2 $\pm$104.4 & \textbf{269.3 $\pm$204.0} \\ 
        
        SAC+QU          & 9530.3 $\pm$ 1670.4          & 114.8$\pm$14.4        & -114.7$\pm$165.4             & 740.6 $\pm$288.8          & 226.0 $\pm$146.8  \\
        \hline
        SAC+MVE         & 9524.8 $\pm$ 1702.6            & N/A                    & 2298.9$\pm$1927.9          & N/A                         & \textbf{1386.7$\pm$1501.2} \\ 
    \textbf{SAC+MVE+BE} & \textbf{9725.2 $\pm$ 929.1}   & N/A                    & \textbf{2382.0$\pm$2185.3} & N/A                         & 1136.4$\pm$1042.9  \\ 
        SAC+MVE+QU      & 9227.4 $\pm$ 1350.3            & N/A                    & 464.2 $\pm$712.5           & N/A                         & 1369.7$\pm$1277.7  \\
        \hline
       \end{tabular}
    \end{table*}
    
\section{Results}
    The primary objective of this experiment is to examine the effect of bounded exploration. Hence, we first contrasted bounded exploration equipped SAC with a vanilla SAC baseline \cite{stable-baselines}. We also evaluated bounded exploration in a model-based setting using MVE. Averaged episodic reward was used to quantify agents' performance. All hyper-parameters were fixed to eliminate ambiguity.  
    The developed agents were tested with Mujoco benchmark \cite{baselines} that contains widely accepted abstractions of robotic tasks with different complexity. For instance, `HalfCheetah-v4' and `Ant-v4' have significant state space. `Hopper-v4' and `Walker-v4' need to learn balancing first. `Swimmer-v4' is considered to be a simple environment with various ways to obtain rewards. All environments' state space and action space are continuous. Algorithms tested with Mujoco can be applied to real-world robots in future. Figure \ref{fig:mesh4} showed a snapshot of the environments. 
    
\begin{figure*}[t]
      \centering
      \begin{subfigure}[b]{0.3\textwidth}
        \includegraphics[width=\textwidth]{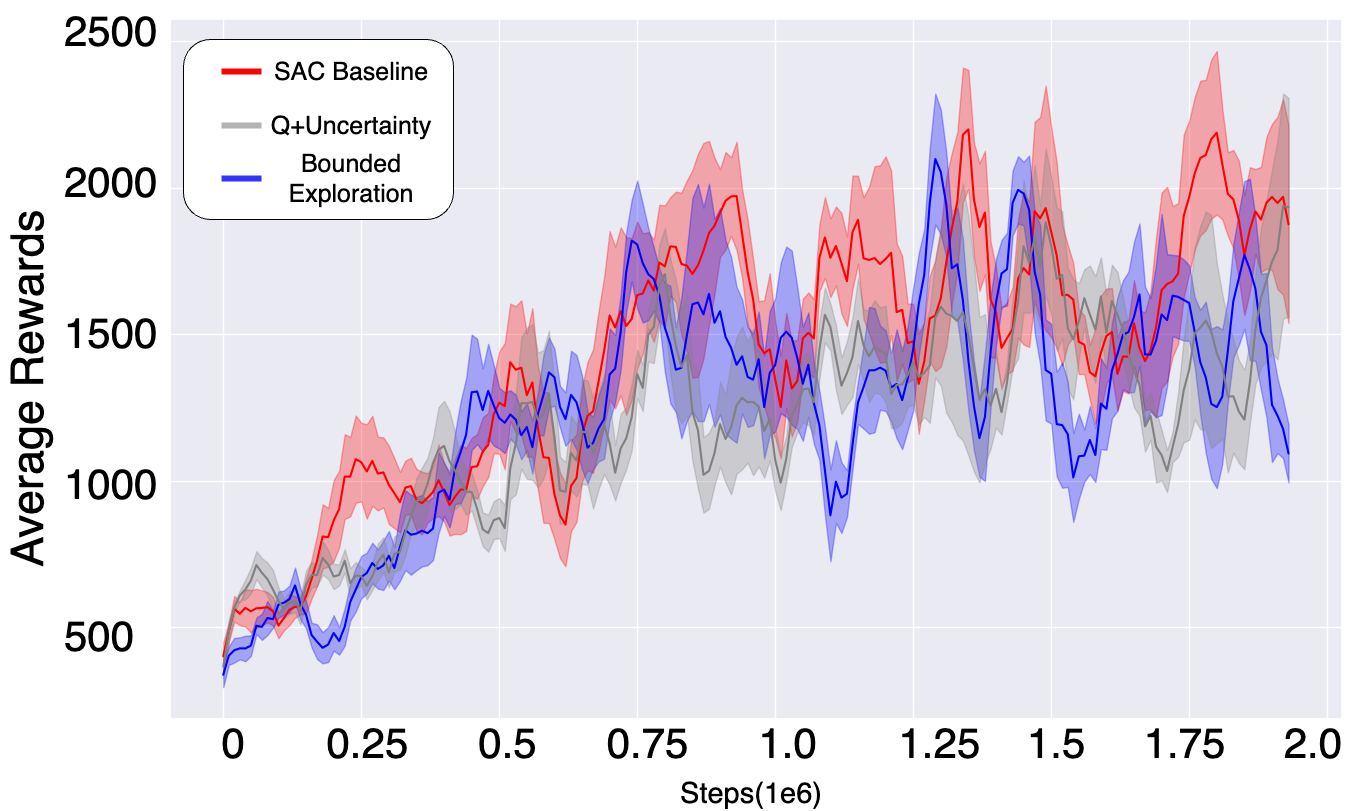}
        \caption{MFRL Hopper-v4}
      \end{subfigure}
      \begin{subfigure}[b]{0.3\textwidth}
        \includegraphics[width=\textwidth]{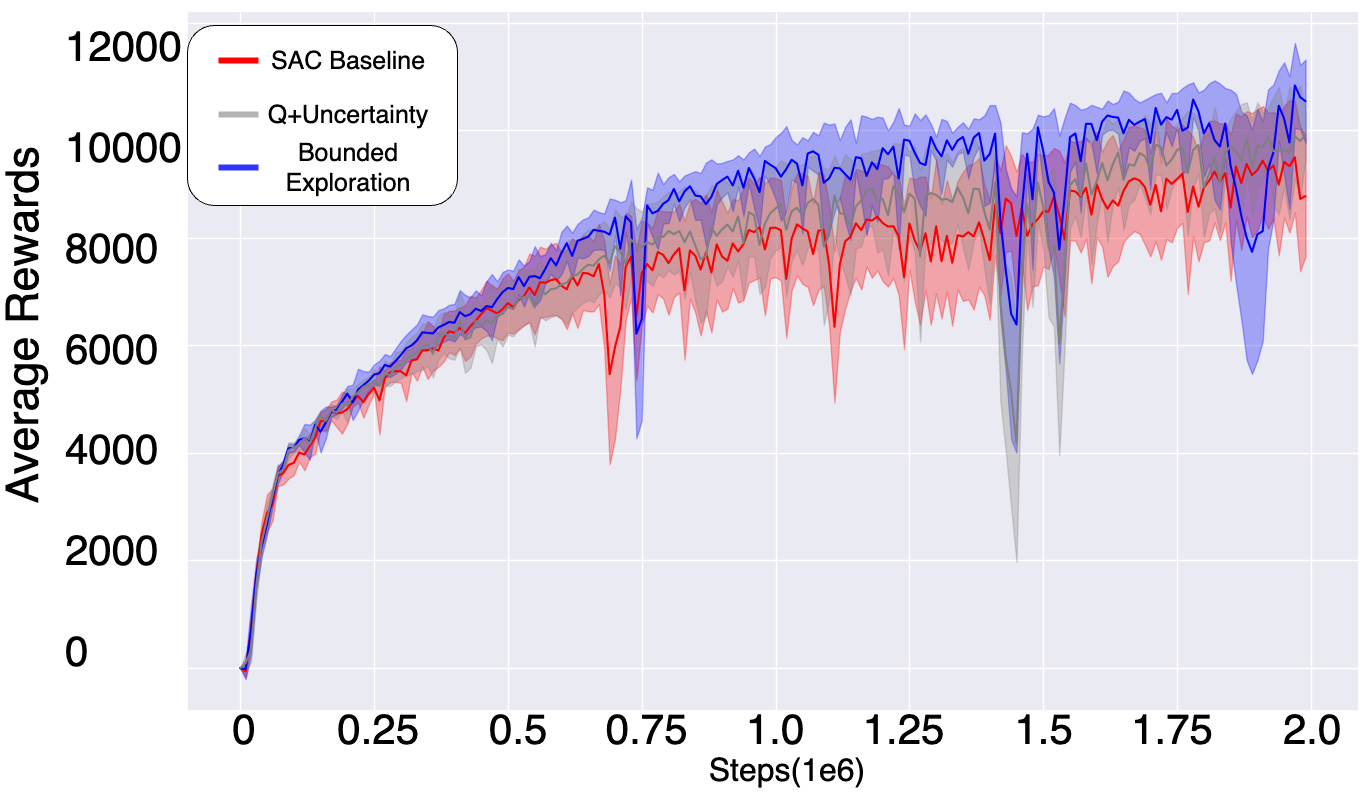}
        \caption{MFRL HalfCheetah-v4}
      \end{subfigure}
      \begin{subfigure}[b]{0.3\textwidth}
        \includegraphics[width=\textwidth]{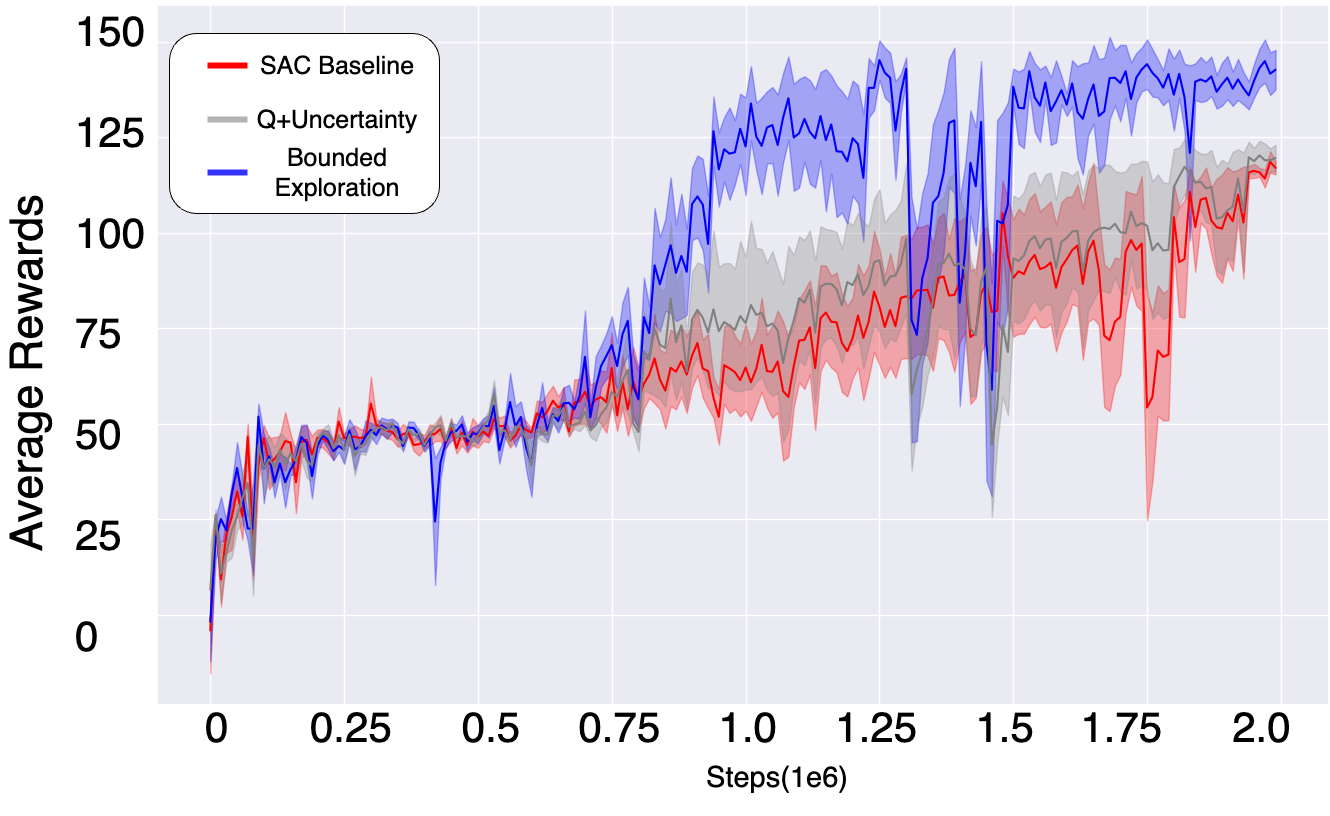}
        \caption{MFRL Swimmer-v4}
      \end{subfigure}
      \hfill
        \vspace{5pt} 
      \begin{subfigure}[b]{0.3\textwidth}
        \includegraphics[width=\textwidth]{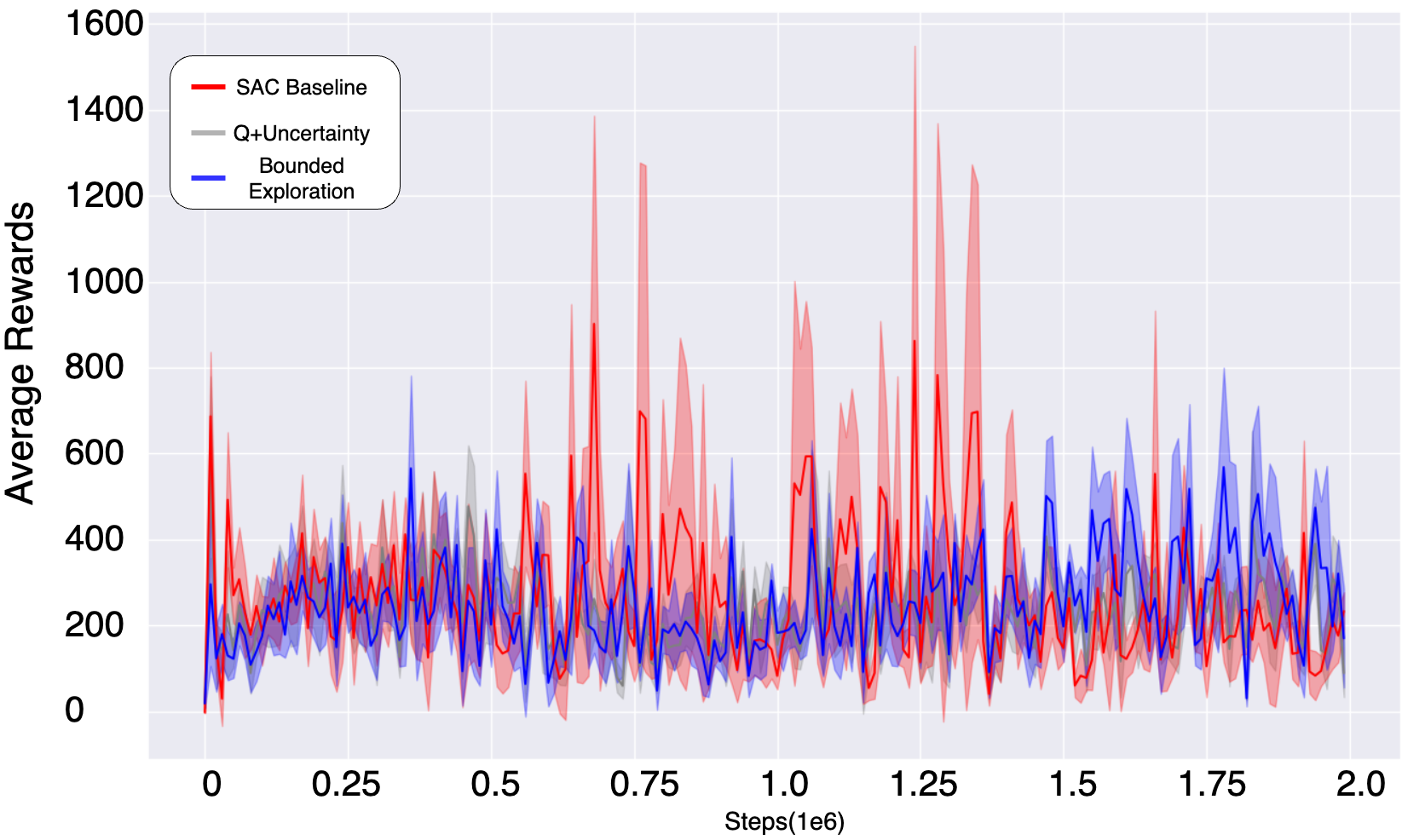}
        \caption{MFRL Walker2d-v4}
      \end{subfigure}
      \begin{subfigure}[b]{0.3\textwidth}
        \includegraphics[width=\textwidth]{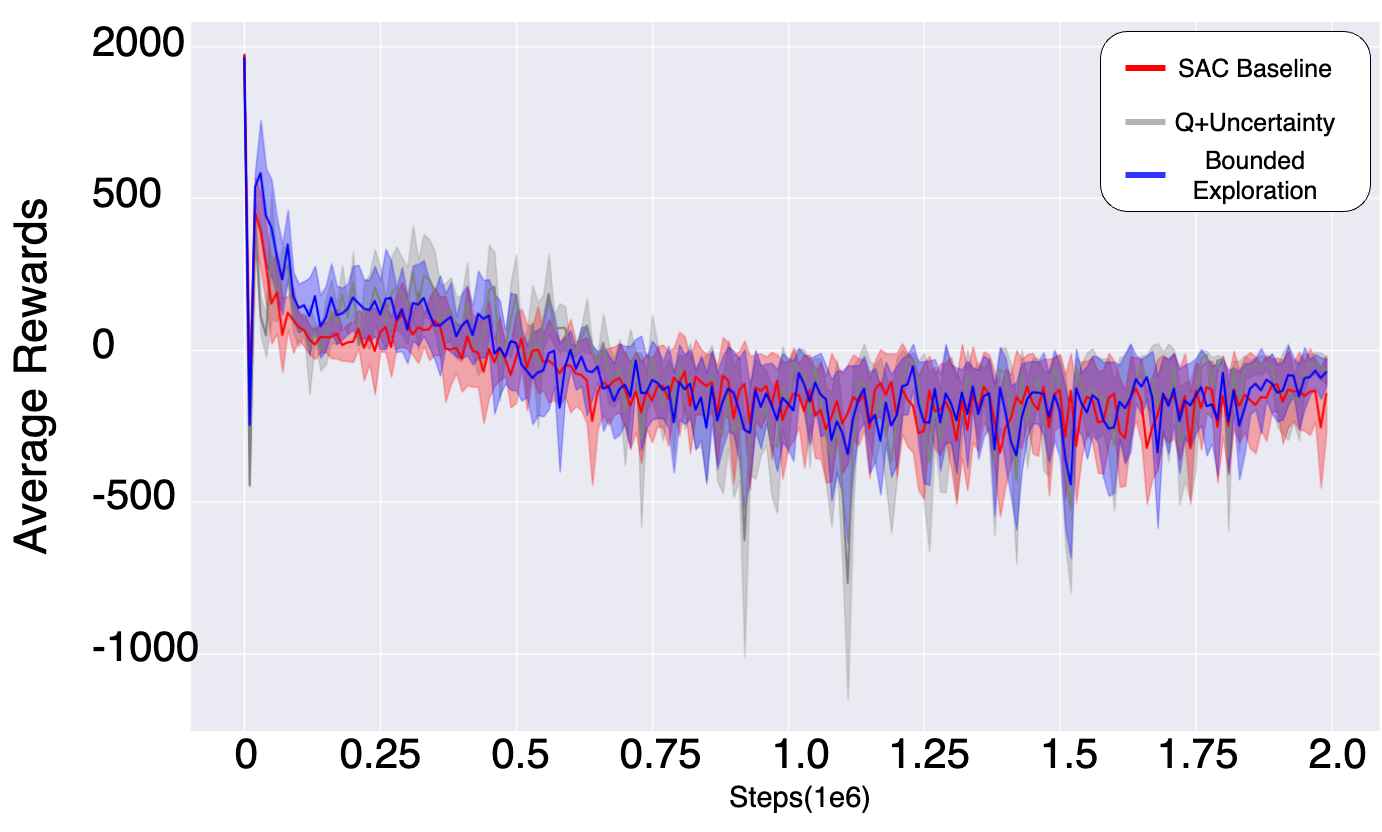}
        \caption{MFRL Ant-v4}
      \end{subfigure}
      \caption{Averaged rewards received by model-free SAC with/without bounded exploration.}
      \label{fig:mesh5}
    \end{figure*}

\begin{figure*}[t]
      \centering
      \begin{subfigure}[b]{0.3\textwidth}
        \includegraphics[width=\textwidth]{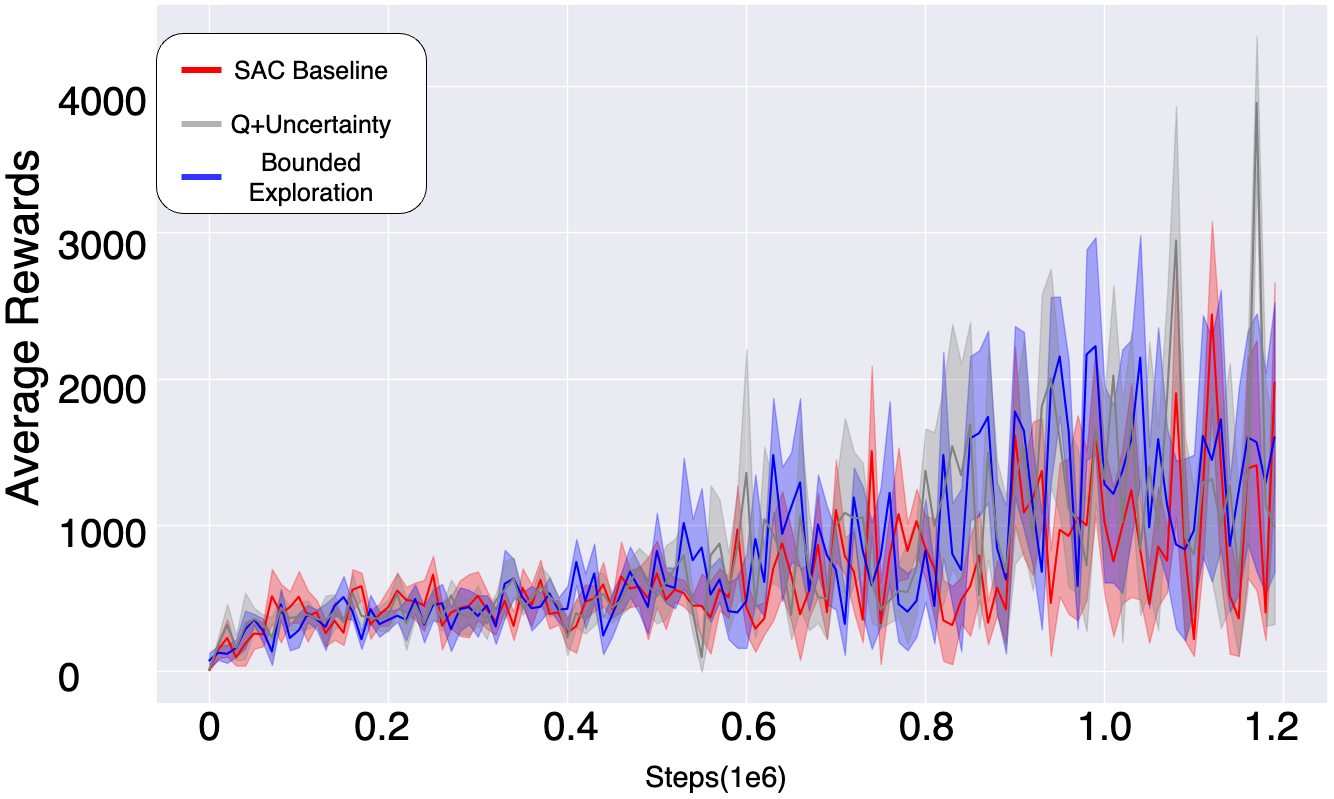}
        \caption{MBRL Walker2d-v4}
      \end{subfigure}
      \begin{subfigure}[b]{0.3\textwidth}
        \includegraphics[width=\textwidth]{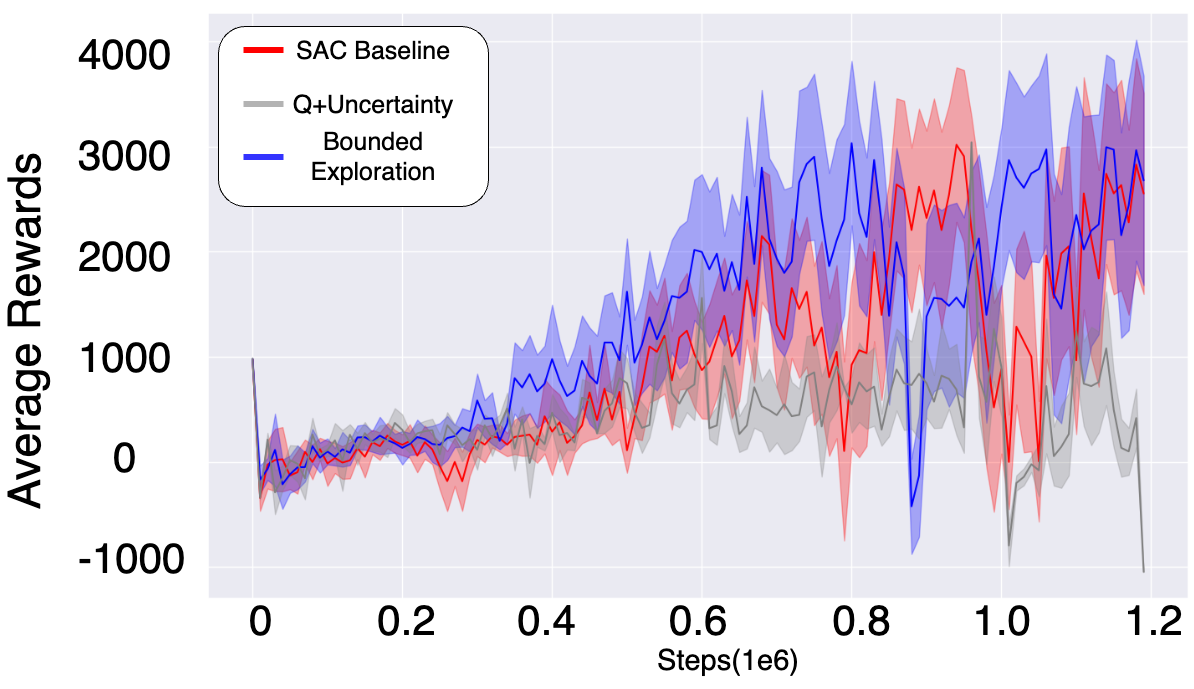}
        \caption{MBRL Ant-v4}
      \end{subfigure}
      \begin{subfigure}[b]{0.3\textwidth}
        \includegraphics[width=\textwidth]{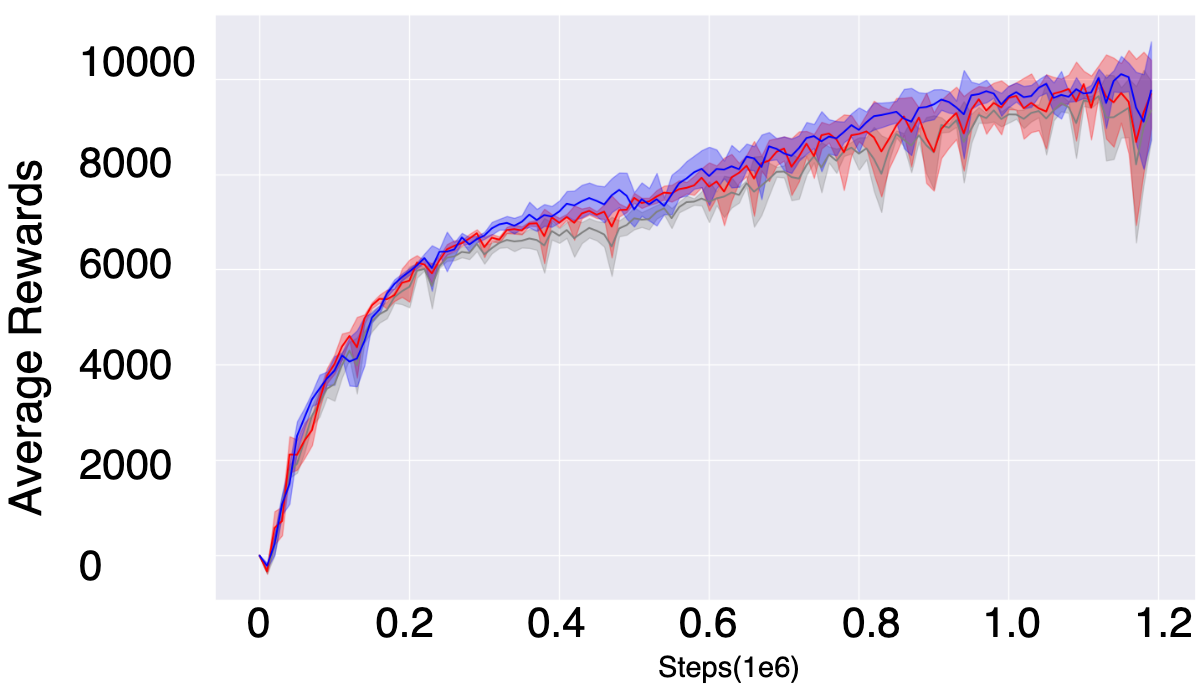}
        \caption{MBRL HalfCheetah-v4}
      \end{subfigure}
      \hfill
      \caption{Averaged rewards received by model-based (Horizon=2) extension of SAC with/without bounded exploration.}
      \label{fig:SAC-results}
      \label{fig:mesh6}
    \end{figure*}

    We plotted the collected evaluation data of an agent during training with (Blue) or without (Red) the bounded exploration part in Figure \ref{fig:mesh5} and Figure \ref{fig:mesh6}. Each environment had been run at least $3$ times to get the mean and variance of received rewards. Each run of an environment consisted of $2 \times 10^6$ training steps for model-free algorithms. It was reduced to $1.2 \times 10^6$ training steps for model-based algorithms since model-based algorithms converged earlier. Based on experimental trial-and-error, we updated every model inside an agent $G=10$ times per environment step. All random seeds (Numpy and Pytorch) were constantly changed at every reset. An agent's performance was regularly recorded at every $1 \times 10^4$ training step. We averaged the reward that the agent received in 10 episodes. For the `Hopper-v4', the collected rewards were averaged with a sliding window (window size $= 10$) to show a clear trend.  

    As Figure \ref{fig:mesh5} and Table \ref{tab:performance-all} demonstrate, SAC with bounded exploration received a higher average reward with fewer episodes in `HalfCheetah-v4' and `Swimmer-v4'. It took less transition data and achieved a higher overall reward. A remarkable performance gain from the "Swimmer-v4" provided compelling evidence that extra exploration with world models' uncertainty in SAC allows an agent to achieve higher rewards in a shorter time. The variance of the received rewards was also plotted at each episode. Unfortunately, in `Hopper-v4', the bounded exploration agent's performance was lower than a vanilla SAC. All three algorithms' performance was volatile and nearly indistinguishable. A possible explanation is that achieving a high reward in `Hopper-v4' can only follow specific patterns. For instance, it needed to stand up first. An agent should be trained to follow such patterns rather than exploring alternatives. Rewards in the other two environments, `HalfCheetah-v4' and `Swimmer-v4', were awarded as long as the agent moved forward.
    
    
    We further studied the `Walker2d-v4' and `Ant-v4' in a model-based setting since MFRL agents did not perform well in these two environments. Regarding MBRL evaluation, as Figure \ref{fig:mesh6} illustrates, the influence of bounded exploration on the MVE was limited. Although the enhancement of bounded exploration was insignificant in achieving higher rewards, an increase in learning speed in `Ant-v4' can be observed. It supported the hypothesis in this paper. A bounded exploration agent can converge with fewer episodes and transitions. Like the `Hopper-v4', `Walker2d-v4' is cumbersome to control. Thus, it was challenging for a bounded exploration agent to learn. 
     
    We attempted the method suggested in \cite{yao2021sample}'s work and plotted it in grey in all resulting graphs. An action with the highest sum of Q-value and variance was selected to interact with the environment at each time step. The results received in the present research are consistent with what they suggested. Both methods yielded a performance gain.   

\section{Discussion}

    Bounded exploration offered an alternative approach to integrate two exploration strategies. The experiment outcomes revealed that bounded exploration can improve data efficiency. It achieved the highest score in $6/8$ experiments. Such learning speed improvement is noteworthy in MFRL. However, a bounded exploration agent did struggle to learn complicated environments. Its performance does not generalize well across environments.   
    
    The evaluation of MBRL with bounded exploration suggested that the improvement is marginal for MBRL. The reason for this result is not completely clear. We are collecting more data. A hypothesis is that we used a random model to make predictions in the value expansion part in MVE. It may already take advantage of the uncertainty, like the work in \cite{chua2018deep}. Using a randomly picked world model in MVE may already have introduced diverse synthetic transitions. Therefore, actively collecting uncertain transitions may have less impact. Further theoretical analysis is required to determine how the uncertainty of world models can affect an agent. 

    Some limitations need to be addressed in this research. Bounded exploration assumes that each dimension of the state is equally important. It could be untrue in some cases. Given that this was only a preliminary attempt, the experiments were limited to simulations with low-dimension states. We are also aware that there are more methods to measure model discrepancy. Our future work has scheduled an investigation of using each of them. 
    
    Regarding implementation, we could not find a simple way to compute uncertainty in the action space without sampling actions for now. A possible solution could be training an inverse dynamic model that inputs $s_t$ and $s_{t+1}$ and output $a_t$, as suggested by Pathak \cite{pathak2017curiosity}.
    
\section{Conclusion and Future Work}

    This paper has presented a novel strategy combining soft RL and intrinsic motivation exploration. In general, the results of this study indicate that jointly deciding an output action based on the agent's policy and intrinsic exploration can benefit data efficiency. The primary concern comes from utilising bounded exploration for model-based reinforcement learning algorithms, especially in complex environments. Further theoretical analysis is expected. Besides, the number of environments we evaluated could be improved. Different intrinsic motivations can be applied. Bounded exploration has meaningful potential in real-world applications. It is suitable for scenarios where intrinsic motivation exploration is needed, but the reward function has a specific meaning and shall not be changed.    

    
\bibliographystyle{named}
\bibliography{bibliography}
\end{document}